\documentclass{article}
\usepackage{ijcai22}

\usepackage{times}
\usepackage{soul}
\usepackage{url}
\usepackage[hidelinks]{hyperref}
\usepackage[utf8]{inputenc}
\usepackage[small]{caption}
\usepackage{graphicx}
\usepackage{amsmath}
\usepackage{amsthm}
\usepackage{booktabs}
\usepackage{algorithm}
\usepackage{algorithmic}
\usepackage{bigstrut}

\title{SSMI: How to Make Objects of Interest Disappear without Accessing Object Detectors?}

\author{
Hui Xia\and
Rui Zhang\and
Zi Kang\and
Shuliang Jiang
\affiliations
College of Computer Science and Technology, Ocean University of China, Qingdao 266100, China\\
\emails
\{zhang\_rui0504)\}@163.com
}

\begin{document}

\maketitle

\begin{abstract}
Most black-box adversarial attack schemes for object detectors mainly face two shortcomings: requiring access to the target model and generating inefficient adversarial examples (failing to make objects disappear in large numbers). To overcome these shortcomings, we propose a black-box adversarial attack scheme based on semantic segmentation and model inversion (SSMI). We first locate the position of the target object using semantic segmentation techniques. Next, we design a neighborhood background pixel replacement to replace the target region pixels with background pixels to ensure that the pixel modifications are not easily detected by human vision. Finally, we reconstruct a machine-recognizable example and use the mask matrix to select pixels in the reconstructed example to modify the benign image to generate an adversarial example. Detailed experimental results show that SSMI can generate efficient adversarial examples to evade human-eye perception and make objects of interest disappear. And more importantly, SSMI outperforms existing same kinds of attacks. The maximum increase in new and disappearing labels is 16\%, and the maximum decrease in mAP metrics for object detection is 36\%.
\end{abstract}

\section{Introduction}

Object detectors demonstrate excellent performance in robot navigation, intelligent video surveillance, industrial inspection, and aerospace, but they also inherit the vulnerability of deep neural networks to adversarial attacks. The adversarial attack generates adversarial examples by adding designed perturbations to input images. The human visual system can barely perceive the differences between adversarial examples and input images. However, testing the adversarial examples on the object detector causes the object detector to fail to recognize objects. The presence of adversarial examples exposes a severe security risk in the application of object detectors. For example, an attacker can easily tamper with the face to induce the object detector to recognize false information; for autonomous driving systems, an attacker can make the object detector fail to identify a stop sign. The study of adversarial attacks helps to understand the mechanism of deep neural networks and helps to stimulate the generation of more powerful defense mechanisms.

\begin{figure}[t]
    \centering            	
    \includegraphics[width=8cm]{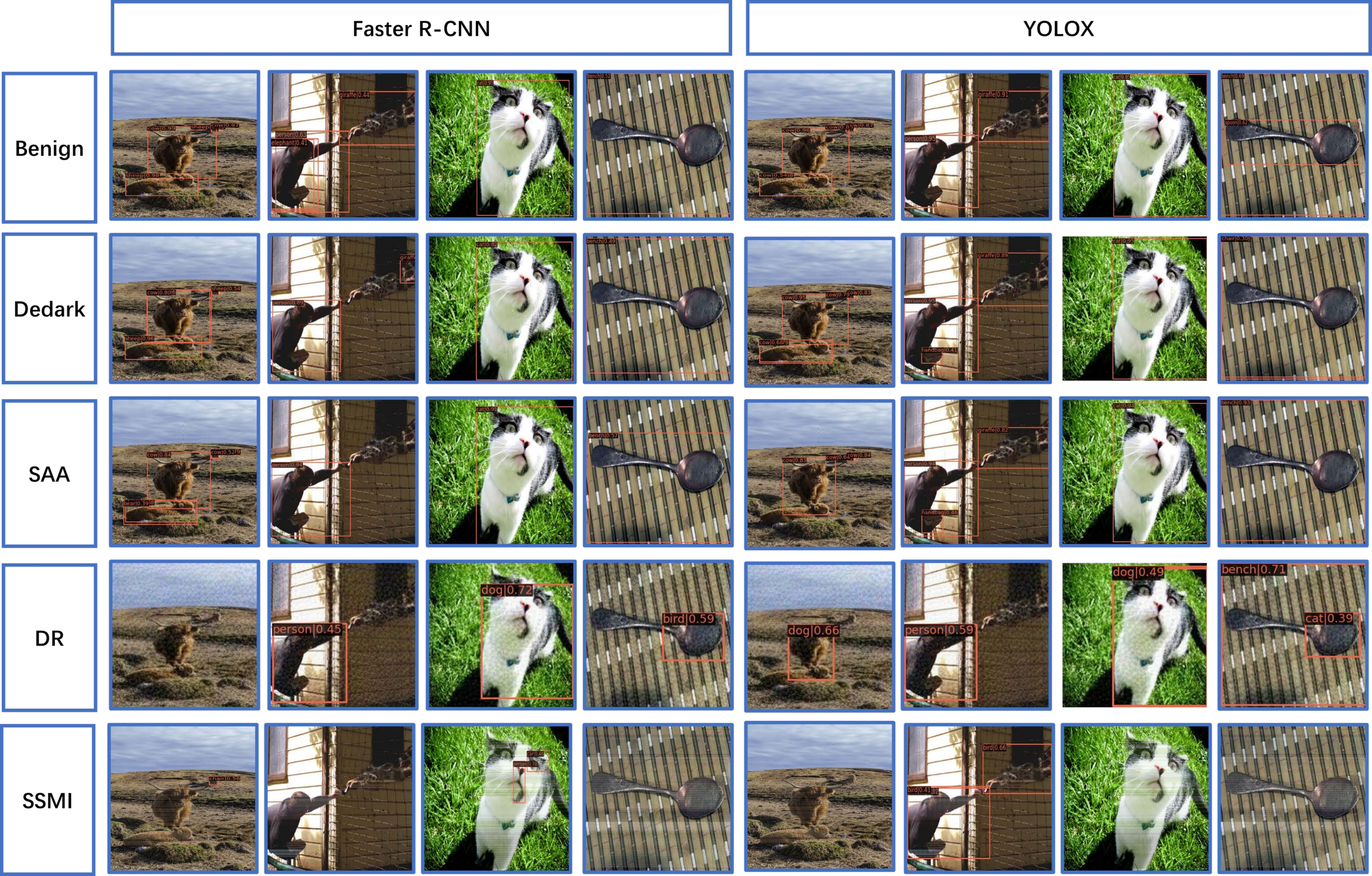}        	
    \caption{Object detection results.}
\end{figure}

Deep neural network-based object detectors can be divided into two categories: candidate region-based (two-stage) and regression-based (one-stage). The major difference between the two is that the former needs to generate candidate bounding boxes with the aid of sub-network ground, while the latter can generate candidate bounding boxes directly on the feature map. The candidate region-based target detection algorithm has an advantage in accuracy, while the regression-based target detection algorithm has an advantage in efficiency. The adversarial attack methods can be classified into two categories: white-box attacks and black-box attacks based on the amount of information inside the object detector that the attacker obtains when launching the attack. White-box attacks generate adversarial examples with high attack success rates by accessing the internal information of the object detector to optimize the objective function, but this leads to difficulties in applying white-box attacks in real scenarios. Black-box attacks require only a small amount of information about the access to the object detector to construct efficient adversarial examples, and thus black-box attacks are more likely to be applied in reality. Black-box adversarial attack methods often construct adversarial examples by adding global perturbations or local perturbations to the input image by accessing a specific object detector. Adding global perturbation to the input image can generate efficient adversarial examples, but it causes a large magnitude of perturbation to the image and large image distortion (as shown in Figure 1). Adding local perturbation to the input image can reduce the distortion of the image and affect the performance of the object detector, but this method still relies on the specific object detector and cannot generate efficient adversarial examples, i.e., it cannot make a large number of objects disappear (as shown in the adversarial examples generated by SAA~\cite{bao2020sparse} and Dedark~\cite{zhao2020object} in Figure 1).

To address the above problems, we propose a pioneering approach, black-box adversarial attack scheme based on semantic segmentation and model inversion, which adds local perturbation to the input image, does not access the target model, does not depend on a specific object detector, and is effective for both candidate region-based (two-stage) and regression-based (one-stage) types of object detectors. The specific contributions are as follows:

(1) To avoid accessing the object detector, we only perturb the regions where the objects exist. That is, we use a Fully Convolutional Neural Network (FCN) to paint the object of interest in a specified color, and then obtain information about the location of the region of the object of interest-based on the color of the object.

(2) we first locate the position of the target object using semantic segmentation techniques. Next, we design a neighborhood background pixel replacement to replace the target region pixels with background pixels to ensure that the pixel modifications are not easily detected by human vision. Finally, we reconstruct a machine-recognizable example and use the mask matrix to select pixels in the reconstructed example to modify the benign image to generate an adversarial example.

(3) To verify the effectiveness of the proposed scheme, four attack schemes, including Dedark, SAA, DR, and SSMI (Our), are used to attack four object detectors, namely Faster R-CNN, YOLOX, Deformable DETR, and CentripetalNet, with the help of six metrics, including the number of Bounding boxes, the number of new labels and the number of disappearing labels. The results show that for Faster R-CNN and YOLOX, SSMI performs comparably with SAA and Dedark. For Deformable DETR and CentripetalNet, SSMI outperforms SAA and Dedark.

\section{Related Work}
This section outlines adversarial attack methods for object detectors in terms of both white-box attacks and black-box attacks.
\subsection{White-box Attacks}
Classical white-box attacks often construct adversarial examples with high attack success rates by optimizing the loss function of the object detector. Xie \emph{et al.}~\cite{xie2017adversarial} densely selected the set of pixel points or the set of target candidate frames to optimize the loss function of the target model and generated efficient adversarial examples, but the method required several iterations of querying the target model to generate efficient adversarial examples. Chen \emph{et al.}~\cite{chen2018shapeshifter} used the Expectation over Transformation technique to detect the robustness of the target detection model for adversarial example attacks, but the method generates poorly migrating adversarial examples and is not suitable for black-box attacks. Liu \emph{et al.}~\cite{liu2018dpatch} proposed an adversarial patches-based attack, Dpatch, which simultaneously optimizes the location of the bounding box and the category objective to invalidate the object detector decision. Similar to Dpatch, Bpatch~\cite{li2018exploring} explores the vulnerability of the Single Shot Module by adding imperceptible patches to the background to corrupt the object detector. However, because patches are modified without additional restrictions, they are usually perceptible to the human eye.

\subsection{Black-box attacks}
The Black-box attack generates high-level adversarial examples by accessing only the input and output of the object detector. Wei \emph{et al.} proposed Unified and Efficient Adversary (UEA), which is based on the Generative Adversarial Network (GAN) framework and combines high-level class loss and low-level feature loss to jointly train an adversarial example generator, greatly reducing the computational cost of constructing adversarial examples~\cite{wei2018transferable}. However, this method is difficult to train and does not have a large improvement in effectiveness compared to white-box attacks. Lu \emph{et al.} reduced the 'dispersion' constraint of the internal feature map when constructing the adversarial examples so that the generated adversarial examples can be effective in many types of target detection tasks, overcoming the shortcomings of existing attack methods that are only specific to task-specific loss functions and target models~\cite{lu2020enhancing}. Wu \emph{et al.} proposed a novel and effective method, called G-UAP, which directly misleads the Region Proposal Network of the detector to mistake the objects as foreground and generate a generic adversarial perturbation~\cite{wu2019g}. Liao \emph{et al.} proposed the first adversarial attack against an unanchored object detection model and used high-level semantic information to generate effective adversarial examples~\cite{liao2020category}, but such methods require a large number of visits to the object detector and cannot effectively attack object detectors with restricted access.

\section{System model}

\subsection{Problem Definition}

The adversarial attack against the target detector is to add an adversarial perturbation to the benign image to fool the target detector. Most previous studies update the adversarial perturbation with details of the target model (e.g., the gradient of the target detector or substitute detector) or input-output pair to make the target detector misidentifies the target object. The objective function is,

 \begin{equation}
 \begin{array}{l}
 \forall {t_n},\arg \mathop {\max } D({\bf{X}} + {\bf{R}}) \ne {l_n}, \\
 s.t.,{\rm{ }}||{\bf{R}}|{|_p} < \eta, \\
 \end{array}
 \end{equation}

 \textbf{X} is the benign image. ${l_n}$ is the true label of the target object. \textbf{R} is a well-designed adversarial perturbation. $|| \cdot |{|_p}$ is the \emph{p} norm, and $\eta$ is used to limit the magnitude of the adversarial perturbation.

Unlike previous studies, we try to design an attack that can personalize the selection of individual or multiple target objects that disappear in benign images. Meanwhile, we update the adversarial perturbation without the detail of the target detector to generate adversarial examples. For this purpose, we replace the target region pixels with background pixels and then add pixels of other labels in the benign image to improve the attack ability of the adversarial example. Our objective function is,

\begin{equation}
\begin{array}{l}
 \arg \mathop {\max }\limits_c D(T({{\bf{X}}_{{t_n}}}) + {{\bf{R}}_{{t_{{n^{'}}}}}}) \ne {l_n}, \\
 s.t.,{\rm{ }}\frac{{{\rm{C(}}T({{\bf{X}}_{{t_n}}}))}}{{{\rm{C(}}{{\bf{X}}_{{t_n}}})}} < \varepsilon ,||{{\bf{R}}_{{t_{{n^{'}}}}}}|{|_2} < \eta , \\
 \end{array}
\end{equation}

${t_n}$ is the target object. $t_n^{'}$ is the other target object. $T({{\bf{X}}_{{t_n}}})$ is the replacement operation of non-target region pixels to target region pixels. ${\rm{ C(}} \cdot )$ is used to count the number of pixels. $\varepsilon$ restricts the ratio of replacement pixels, $\varepsilon  \ge \frac{1}{4}$. ${{\bf{R}}_{{t_{{n^{'}}}}}}$ denotes the target object pixels. $\eta$ is used to restrict the change of other target pixels to the target region. An effective adversarial example needs to satisfy two elements: (\emph{i}) the adversarial example generated by adding the adversarial perturbation to the benign image does not affect human visual perception and is not easily perceived by the human eye. (\emph{ii}) The generated adversarial examples have the ability to fool the target model. That is, the target detector is not capable of identifying the target object that we have personalized to choose to attack. To satisfy the above two elements,
we first locate the position of the target object using semantic segmentation techniques. Next, we design a neighborhood background pixel replacement to replace the target region pixels with background pixels to ensure that the pixel modifications are not easily detected by human vision. Finally, we reconstruct a machine-recognizable example and use the mask matrix to select pixels in the reconstructed example to modify the benign image to generate an adversarial example.

\subsection{Object Location}
We try various target location methods to determine the location of freely selected target objects, such as traditional target location and DNN-based target detection methods. However, these methods can only obtain information about the bounding box of the target object. They cannot accurately and flexibly describe the contour curve of the target object, which is not conducive to our accurate localization of the target object. Image semantic segmentation can label each pixel of a benign image, and pixels with the same label are filled with the same color. Therefore, we can precisely locate the region of the target object to be attacked based on the color that different target objects are filled.

All the annotation information of the COCO 2017 dataset is saved in JSON files, from which we can extract the semantic segmentation of the mask. Besides, we can also use semantic segmentation techniques to segment the semantics of benign images. For example, FCN~\cite{long2015fully} semantically segments \textbf{X} and then fills out the color information of the object based on the semantic segmentation to obtain the region location of the object. Let ${x_{i, j}}$ be the vector of the input image\textbf{ X} at position (\emph{i},\emph{j}), and the output ${y_{i,j}}$ of the next layer after one layer of the convolutional neural network is,
\begin{equation}
{y_{ij}} = {f_{ks}}({\{ {{\bf{X}}_{si + \delta i,sj + \delta j}}\} _{0 \le \delta i,\delta i \le k}}),
\end{equation}
\emph{k} is the size of the convolution kernel, \emph{s} is the step size, $\delta$ is the offset, and ${f_{ks}}( \cdot )$ is the type of convolution layer. To obtain semantic information of the image with higher accuracy, Long \emph{et al.} used the strategy of hopping connections, multiple upsampling then combining the high-level information, and finally upsampling to the input image size. The upsampling is actually a bilinear interpolation operation, i.e., one linear interpolation in the \emph{a} and \emph{b} directions, respectively. The values of the four points of the function \emph{f} are known to be ${Q_{11}}  =  \left( {{a_1}, {b_1}} \right),{Q_{12}}  =  \left( {{a_1}, {b_2}} \right), {Q_{21}}  =  \left( {{a_2}, {b_1}} \right)$, and ${Q_{22}}  =  \left( {{a_2}, {b_2}} \right)$. A linear interpolation in the direction of \emph{a} (horizontal coordinate) yields that
\begin{equation}
  f(a,{b_1}) \approx \frac{{{a_2} - a}}{{{a_2} - {a_1}}}f({Q_{11}}) + \frac{{a - {a_1}}}{{{a_2} - {a_1}}}f({Q_{21}}),
\end{equation}

\begin{equation}
f(a,{b_2}) \approx \frac{{{a_2} - a}}{{{a_2} - {a_1}}}f({Q_{12}}) + \frac{{a - {a_1}}}{{{a_2} - {a_1}}}f({Q_{22}}),
\end{equation}
Then the linear interpolation in the direction of \emph{b} (vertical coordinate) is,
\begin{equation}
\begin{aligned}
f\left( {a,{\rm{ }}b} \right)&\approx \frac{{{b_2} - b}}{{{b_2} - {b_1}}}f(a,{b_1}) + \frac{{b - {b_1}}}{{{b_2} - {b_1}}}f(a,{b_2})\\
&= \frac{{[\begin{array}{*{20}{c}}
   {{a_2} - a} & {a - {a_1}}  \\
\end{array}][\begin{array}{*{20}{c}}
   {f({Q_{11}})} & {f({Q_{12}})}  \\
   {f({Q_{21}})} & {f({Q_{22}})}  \\
\end{array}][\begin{array}{*{20}{c}}
   {{b_2} - b}  \\
   {b - {b_1}}  \\
\end{array}]}}{{({a_2} - {a_1})({b_2} - {b_1})}}
\end{aligned}
\end{equation}
Finally, the pixels of the object of interest are painted in the specified color. For example, all the pixels of the dog are filled with blue color, and the location information of the target dog is obtained from the root blue pixel value.

\subsection{Adversarial Examples Generation}
\subsubsection{Neighbor Background Pixel Replacement}
The neighborhood background pixel replacement to replace the target region pixels with background pixels, i.e., the input image is perturbed with the neighboring pixels of the target region by step.

\subsubsection{Label Fixed Sample Reconstruction}
We reconstruct machine-recognizable samples in the following two steps: (a) train a well-performing classifier $C({\bf{X}};\theta)$, (b) fix the output label (the true label of the object) using a model inversion technique to update the gradient information of the image based on the target loss function, and reconstruct the machine-recognizable sample ${{\bf{S}}_{{t_n}}}$. It is worth noting that we do not require the reconstructed sample to be visually identical to the input image, but only to ensure that the machine recognizes it as the label of the object of interest (only machine recognizable), and the objective function of the reconstructed sample is,
\begin{equation}
loss = {\bf{1}} - C{({{\bf{S}}_{{t_n}}};\theta )_{target}},
\end{equation}
${\bf{S}}_{{t_n}}$ is set to an all-0 matrix of the same size as \textbf{X}. $C{({{\bf{R}}_{{t_n}}};\theta )_{target}}$ is the probability of label of the object of interest. For example, when reconstructing an image classified as a dog, $C{({{\bf{S}}_{{t_n}}})_{target}}$ is the probability that the classifier classifies a dog. We use the gradient descent with momentum to update ${\bf{S}}_{{t_n}}$.
\begin{equation}
{\bf{V}} = {\lambda _1}*{\nabla _{{{\bf{R}}_{{t_n}}}}}loss + {\lambda _2}*{\bf{V}},
\end{equation}
\begin{equation}
TV({\bf{V}}) = \sum\limits_{i,j} {|{{\bf{V}}_{i + 1,j}} - {{\bf{V}}_{i,j}}| + } |{{\bf{V}}_{i,j + 1}} - {{\bf{V}}_{i,j}}|,
\end{equation}
\begin{equation}
{\bf{S}}_{{t_n}}^* = (1 - \alpha )*{{\bf{S}}_{{t_n}}} + \beta *TV({\bf{V}}).
\end{equation}
$TV({\bf{V}})$ is total variation loss which is designed to make the image more smooth. Initially, \textbf{V} is a 0 matrix of the same size as ${\bf{S}}_{{t_n}}$. ${\lambda _1},{\lambda _2},\alpha,$ and $\beta$ are scalar weights, where ${\lambda _1} + {\lambda _2} = 1,{\lambda _1} = 0.5,\alpha  = 0.1,\beta  = 0.01$.
\subsubsection{Generation Adversarial Example}
We use mask matrices $M_{n \times m}^1$ and $M_{n \times m}^2$ with different step sizes to find significant features in the reconstruction sample as local perturbations. Take input image ${{\bf{X}}_{w \times h \times c}}$, \emph{w} is the width of \textbf{X}, \emph{h} is the height of \textbf{X}, and \emph{c} is the number of channels of \textbf{X}, $c=3, n = 1,m = 10$, for example, The perturbations ${{\bf{R}}_{{t_n}}}$ in the \emph{a}-direction and \emph{b}-direction are,
\begin{equation}
{{\bf{R}}_{{t_n}}}[:,s:s + m,:] = {\bf{S}}_{{t_n}}^*[:,s:s + m,:]
\end{equation}

\begin{equation}
{{\bf{R}}_{{t_n}}}[s:s + n,:,:] = {\bf{S}}_{{t_n}}^*[s:s + n,:,:]
\end{equation}
if the coordinates of the object are $(t,b,l,r)$, \emph{t} is the top position, \emph{b} is the bottom position, \emph{l } is the left boundary, and \emph{r} is the right boundary. The adversarial example ${{\bf{X}}_{adv}}$ is,
\begin{equation}
{{\bf{X}}_{adv}} = {{\bf{X}}_{adv}^{'}}[i + t:b,j + l:r,:] + {\bf{R}}_{{t_n}}^*[:,:,:]
\end{equation}
(\emph{i}, \emph{j}) is the coordinate in the \emph{a}-direction and \emph{b}-direction.

\section{Stimulation Results}

\subsection{Experiment Settings}

To verify the effectiveness of SSMI, six metrics, namely Average Precision (AP), Average Recall (AR), bbox$\_$mAP, the number of Bounding boxes, the number of new labels, and the number of disappearing labels, are used. Four attack schemes, namely Dedark, DR, SAA, and SSMI, are used to attack four object detectors, namely Faster R-CNN~\cite{ren2015faster}, YOLOX~\cite{ge2021yolox}, Deformable DETR~\cite{zhu2020deformable}, and CentripetalNet~\cite{dong2020centripetalnet}.

\begin{table}[t]
  \centering
  \small
  \renewcommand\tabcolsep{4.0pt}

    \begin{tabular}{p{3em}cccccc}
    \hline
    \textbf{Method} & \textbf{Origin} & \textbf{SAA} & \textbf{Dedark} &\textbf{SSMI} & \textbf{DR} & \textbf{SSMI+A} \bigstrut[t]\\
    \hline
    \textbf{FRCNN} & 155714 & \textbf{141656} & 142458 & 146427 & 61790 & 103982 \bigstrut[t]\\
    \textbf{YOLOX} & 176661 & 185818 & 192042 & \textbf{167411} & 131915 & 132193 \\
    \textbf{DDETR} & \textbf{500000} & \textbf{500000} & \textbf{500000} & 500000 & 500000 & 500000 \\
    \textbf{CNet} & 460871 &\textbf{ 461897} & 461666 & 477405 & 484178 & 482402 \bigstrut[b]\\
    \hline
    \end{tabular}%
    \caption{Number of bounding boxes}
  \label{tab:addlabel}%
\end{table}%

\begin{table}[t]
  \centering
  \small

  \begin{tabular}{p{3em}cccccc}
    \hline
    \textbf{Method} & \textbf{SAA} & \textbf{Dedark} & \textbf{SSMI} & \textbf{DR} & \textbf{SSMI+A} \bigstrut\\
    \hline
    \textbf{FRCNN} & 30362 & 26757 & \textbf{45901} & 24882 & \textbf{61781} \bigstrut[t]\\
    \textbf{YOLOX} & \textbf{48967} & \textbf{50571} & 42837 & 61088 & \textbf{82622} \\
    \textbf{DDETR} & 134337 & 113701 & \textbf{193198} & 288916 & \textbf{304701} \\
    \textbf{CNet} & 114407 & 101946 & \textbf{156076} & 244149 & \textbf{299127} \bigstrut[b]\\
    \hline
    \end{tabular}%
    \caption{Number of new labels}
  \label{tab:addlabel}%
\end{table}%

\begin{table}[t]
  \centering
  \small

    \begin{tabular}{p{3em}ccccc}
    \hline
    \textbf{Method} & \textbf{SAA} & \textbf{Dedark} & \textbf{SSMI} & \textbf{DR} & \textbf{SSMI+A} \bigstrut\\
    \hline
    \textbf{FRCNN} & 44420 & 40013 & \textbf{55188} & \textbf{118806} & 113510 \bigstrut[t]\\
    \textbf{YOLOX} & 39810 & 35190 & \textbf{52087} & 105834 & \textbf{127090} \\
    \textbf{DDETR} & 134337 & 113701 & \textbf{193198} & 288916 & \textbf{304701} \\
    \textbf{CNet} & 111381 & 101151 & \textbf{139542} & 220842 & \textbf{277596} \bigstrut[b]\\
    \hline
    \end{tabular}%
    \caption{Number of disappearing labels}
  \label{tab:addlabel}%
\end{table}%

\begin{figure}[t]
    \centering            	
    \includegraphics[width=7cm]{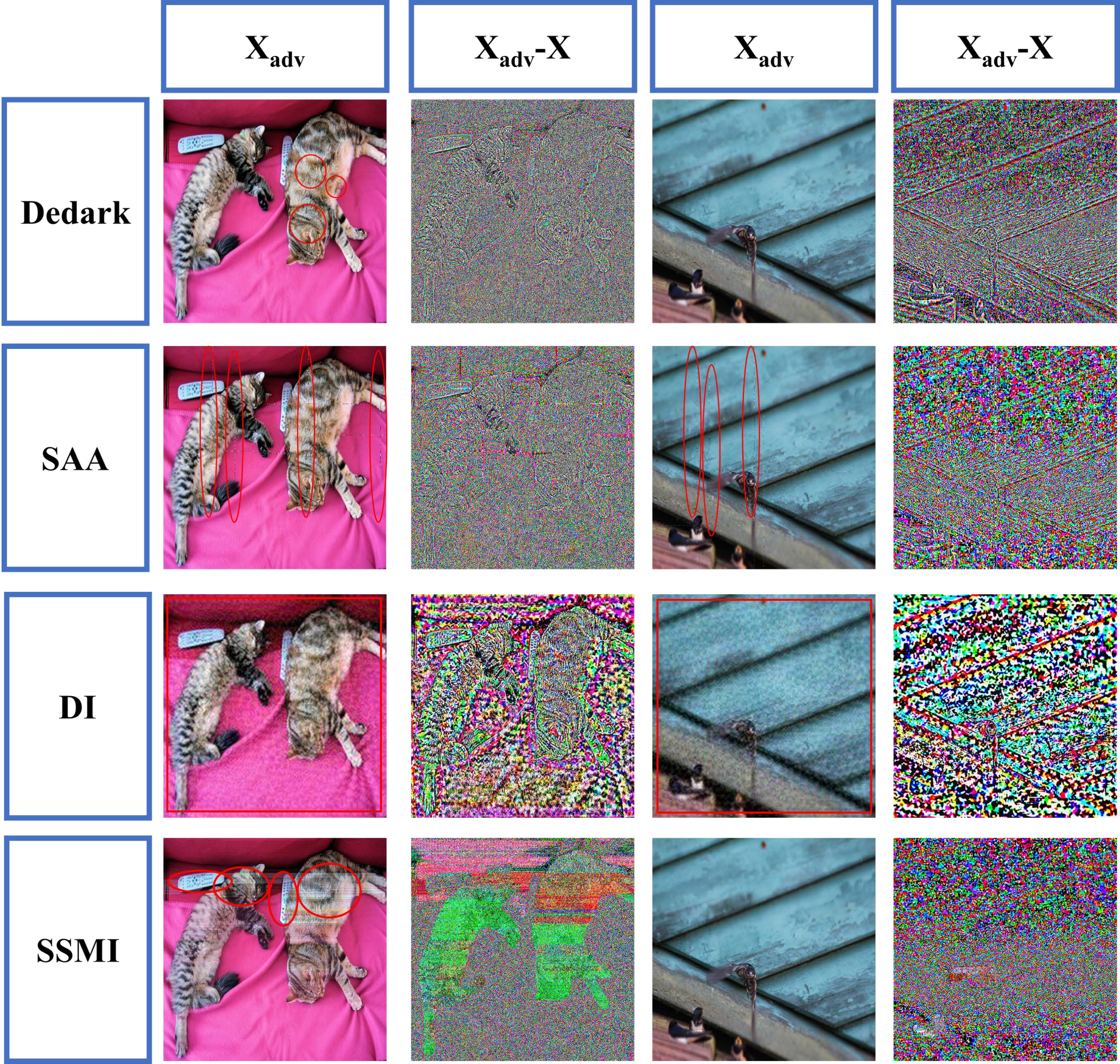}        	
    \caption{Object detection results.}
\end{figure}

\subsection{Perturbation Region Comparison}
Figure 2 shows the differences between the adversarial examples generated by the four attack schemes of Dedark, SAA, DR, and SSMI and the input image (Begin). ${{\bf{X}}_{adv}}$ is the adversarial example, ${{\bf{X}}_{adv}} - {\bf{X}}$ is the difference between the adversarial example and begin image, and the red marker indicates the location of the perturbation. Analysis of this figure from the perspective of the perturbed region shows that the DR-generated adversarial example differs the most from the input image, which is due to the fact that DR adds perturbation to the entire input image. Analyzed from the perspective of human eye perception, the SSMI perturbed area is larger than that of SAA and Dedark, but it can meet the evasion of human eye easy perception.

\subsection{Comparison of Object Detection Results}
Figure 1 shows the results of the four attack schemes, Dedark, DR, SAA, and SSMI, for generating adversarial examples of images containing single target objects against the two target detectors, Faster R-CNN and YOLOX. From this figure, it is clear that the adversarial examples generated by the four attack schemes, Dedark, SAA, DR, and SSMI, can all affect the detection results of the target detectors to different degrees. For example, Dedark misleads the Faster R-CNN to misidentify cow as sheep, narrows the detection region of giraffe, reduces the recognition probability of cat and bench, and misleads the YOLOX to misidentify human hand as a handbag. SAA can mislead the Faster R-CNN to fail to recognize giraffe and narrow the detection region of the bench. DR can mislead Faster R-CNN and YOLOX to fail to recognize cow and giraffe, misidentify cat as dog and spoon as a bird. SSMI misleads Faster R-CNN and YOLOX to fail to recognize cow, cat, and spoon, misleads Faster R-CNN to misidentify cat as person and car, and misleads YOLOX to misidentify person and giraffe as a bird. Only SSMI can make both Faster R-CNNN and YOLOX target detectors effective for four images.

\begin{table}[t]
  \centering
  \small
  \renewcommand\tabcolsep{4.0pt}
    \begin{tabular}{p{3em}cccccc}
    \hline
    \textbf{Method} & \textbf{Origin} & \textbf{SAA} & \textbf{Dedark} &\textbf{SSMI} & \textbf{DR} & \textbf{SSMI+A} \bigstrut[t]\\
    \hline
    \textbf{FRCNN} & 155714 & \textbf{141656} & 142458 & 146427 & 61790 & 103982 \bigstrut[t]\\
    \textbf{YOLOX} & 176661 & 185818 & 192042 & \textbf{167411} & 131915 & 132193 \\
    \textbf{DDETR} & \textbf{500000} & \textbf{500000} & \textbf{500000} & 500000 & 500000 & 500000 \\
    \textbf{CNet} & 460871 &\textbf{ 461897} & 461666 & 477405 & 484178 & 482402 \bigstrut[b]\\
    \hline
    \end{tabular}%
    \caption{Number of bounding boxes}
  \label{tab:addlabel}%
\end{table}%

\begin{table}[t]
  \centering
  \small

  \begin{tabular}{p{3em}cccccc}
    \hline
    \textbf{Method} & \textbf{SAA} & \textbf{Dedark} & \textbf{SSMI} & \textbf{DR} & \textbf{SSMI+A} \bigstrut\\
    \hline
    \textbf{FRCNN} & 30362 & 26757 & \textbf{45901} & 24882 & \textbf{61781} \bigstrut[t]\\
    \textbf{YOLOX} & \textbf{48967} & \textbf{50571} & 42837 & 61088 & \textbf{82622} \\
    \textbf{DDETR} & 134337 & 113701 & \textbf{193198} & 288916 & \textbf{304701} \\
    \textbf{CNet} & 114407 & 101946 & \textbf{156076} & 244149 & \textbf{299127} \bigstrut[b]\\
    \hline
    \end{tabular}%
    \caption{Number of new labels}
  \label{tab:addlabel}%
\end{table}%

\begin{table}[t]
  \centering
  \small

    \begin{tabular}{p{3em}ccccc}
    \hline
    \textbf{Method} & \textbf{SAA} & \textbf{Dedark} & \textbf{SSMI} & \textbf{DR} & \textbf{SSMI+A} \bigstrut\\
    \hline
    \textbf{FRCNN} & 44420 & 40013 & \textbf{55188} & \textbf{118806} & 113510 \bigstrut[t]\\
    \textbf{YOLOX} & 39810 & 35190 & \textbf{52087} & 105834 & \textbf{127090} \\
    \textbf{DDETR} & 134337 & 113701 & \textbf{193198} & 288916 & \textbf{304701} \\
    \textbf{CNet} & 111381 & 101151 & \textbf{139542} & 220842 & \textbf{277596} \bigstrut[b]\\
    \hline
    \end{tabular}%
    \caption{Number of disappearing labels}
  \label{tab:addlabel}%
\end{table}%

\subsubsection{Object disappearance analysis}

Tables 1-Table 3 show the changes in the three metrics of the number of bounding boxes, the number of new labels, and the number of disappearing labels when four targets are detected by simultaneous attacks of Dedark, SAA, DR, and SSMI on Faster R-CNN, YOLOX, Deformable DETR, and CentripetalNet. As can be seen from Fig. 2, both Dedark and SAA schemes perturb the object locally, and DR perturbs the whole image. To ensure fairness, we set SSMI+A to perturb the input image in the same proportion as the DI scheme, i.e., DR perturbs the whole image and SSMI+A perturbs the whole object. As can be seen from Table 1, SSMI has the least number of bounding boxes in the adversarial examples constructed for the Faster R-CNN and CentripetalNet SAA compared to Dedark and SAA, while SSMI has the least number of bounding boxes for YOLOX. This is because the number of added labels in the SSMI-constructed adversarial example is more than the number of added labels in the SAA-constructed adversarial example, as shown in Table 2. It should be noted that for the Deformable DETR, the number of bounding boxes in all three schemes, Dedark, SAA, and SSMI, is the same as the number of bounding boxes in the original image, because the number of new labels and the number of disappearing labels for the Deformable DETR are equal, as shown in Tables 2 and 3.

From Table 2, it can be seen that both SAA and Dedark add more labels than the most for the YOLOX, and SSMI adds the most labels (only 1\% more) only for the three object detectors Faster R-CNN, Deformable DETR, and CentripetalNet, which is because we focus on making the objects disappear rather than adding new labels. As shown in Table 3, SSMI generates more disappearing labels than SAA and Dedark for four object detectors, up to 193196, with 39\% of the labels disappearing. In summary, SSMI is able to make the highest number of disappearing labels in the generated adversarial examples compared with SAA and Dedark, which can satisfy the purpose of making the objects disappear.

Comparing DR and SSMI+A, we can see that the number of new labels and the number of disappearing labels in the adversarial examples generated by SSMI+A against the four object detectors is higher than the number of new labels and the number of vanishing labels in the adversarial examples generated by DR, which further verifies the effectiveness of SSMI.

\subsubsection{Impact of adversarial examples on model performance}

Tables 4-Table 6 show the impact of the SSMI-generated adversarial examples on the Average Precision (AP), Average Recall (AR), and ${\rm{bbox\_mAP}}$ metrics of the four object detectors, Faster R-CNN, YOLOX, Deformable DETR, and CentripetalNet. As can be seen from Table 4, SSMI has the most serious impact on the ${\rm{A}}{{\rm{P}}_{{\rm{50}}}}$ metric of Deformable DETR, followed by Faster R-CNN object detector, with the ${\rm{A}}{{\rm{P}}_{{\rm{50}}}}$ metric dropping by 40\% and 32\%, respectively. As can be seen from Table 5, SSMI has the most serious impact on the ARM metric of Deformable DETR, with the ARM metric dropping by 30\%. As can be seen from Table 6, SSMI has the most serious impact on the\emph{ mAP} metric of YOLOX, which drops 36\%, on the ${\rm{bbox\_mAP}}$ metric of YOLOX object detector, which drops 36\%, on the ${\rm{bbox\_mA}}{{\rm{P}}_{50}}$ metric of Faster R-CNN object detector, which drops 32\%, and ${\rm{mA}}{{\rm{P}}_{\rm{L}}}$ for the Deformable DETR, where the metric decreased by 40\%. In summary, the adversarial examples generated by SSMI are valid for all four object detectors.

\section{Conclusion}
Designing effective adversarial attack methods for object detectors has motivated more robust defense schemes, but previous research on this problem has mainly faced the problem of requiring access to the target model. The pure black-box no-access attack approach proposed in we generates efficient adversarial examples without accessing the object detector and is effective for multiple object detectors. Detailed experimental results verify the effectiveness of SSMI. In future research, we will continue to try to explore novel adversarial example generation methods and ensure that the perturbations added to the original image are not easily perceived by the human eye.

\bibliographystyle{alpha}
\bibliography{ijcai22-arxive}

\appendix

\end{document}